\documentclass{article}

\usepackage{PRIMEarxiv}
\usepackage{multirow}
\usepackage[utf8]{inputenc} 
\usepackage[T1]{fontenc}    
\usepackage{hyperref}       
\usepackage{url}            
\usepackage{booktabs}       
\usepackage{amsfonts}       
\usepackage{nicefrac}       
\usepackage{microtype}      
\usepackage{lipsum}
\usepackage{fancyhdr}       
\usepackage{graphicx}       
\usepackage{cite}
\graphicspath{{media/}}     

\title{Relational Graph Convolutional Networks for Sentiment Analysis}

\author{
  Asal Khosravi, Zahed Rahmati, Ali Vefghi \\
  Department of Mathematics and Computer Science \\
  Amirkabir University of Technology \\
  \texttt{\{a.khsoravi96, zrahmati, alivefghi\}  @aut.ac.ir} \\
  }
  
\begin{document}
\maketitle

\begin{abstract}

  With the growth of textual data across online platforms, sentiment analysis has become crucial for extracting insights from user-generated content. 
  While traditional approaches and deep learning models have shown promise, they cannot often capture complex relationships between entities. 
  In this paper, we propose leveraging Relational Graph Convolutional Networks (RGCNs) for sentiment analysis, which offer interpretability and flexibility by capturing dependencies between data points represented as nodes in a graph. 
  We demonstrate the effectiveness of our approach by using pre-trained language models such as BERT and RoBERTa with  RGCN architecture on product reviews from Amazon and Digikala datasets and evaluating the results. 
  Our experiments highlight the effectiveness of RGCNs in capturing relational information for sentiment analysis tasks.

\end{abstract}


\section{Introduction}

Sentiment analysis, also known as opinion mining, is a fundamental task in natural language processing within the broader domain of text classification which aims to extract valuable insights from various social platforms and extensive online texts, enabling the analysis of people's attitudes across various domains including business, advertising, government, economics, and even political orientations. 
As the utilization of text in online conversations, emails, and user-generated sentiments on the internet about various products, movies, and services continues to rise, there emerges a pressing need for robust mechanisms capable of analyzing and interpreting textual data. 
Text classification involves categorizing text documents into predefined classes or categories based on their content. 
In the case of sentiment analysis, the objective is to classify text documents into categories representing different sentiments or emotions, such as positive, negative, or neutral. 
Researchers have made many efforts in this field, which mainly refer to traditional approaches based on dictionaries, machine learning, and deep learning models. 
Despite achieving impressive results in sentiment analysis, deep learning models can be unexplainable. 
This "black box" nature arises from difficulties in interpreting the model's internal workings, such as the weights assigned to features and the high dimensionality of the feature space itself.

Graph Neural Networks (GNNs) have emerged as a powerful paradigm for analyzing structured data, offering unique advantages in capturing relationships and dependencies between data points represented as nodes in a graph. 
GNNs excel at exploiting the rich relational information inherent in graph structures. 
Unlike traditional Graph neural networks, which usually treat all connections the same, Relational Graph convolutional networks consider the different types of relationships in the graph. 
This allows them to better understand and analyze interconnected data.

Heterogeneous graphs, where nodes and edges can have different types, offer a natural representation of real-world systems with diverse relationships. 
While successful for text classification, traditional GCNs treat all relationships in text graphs as homogeneous. 
By treating all relationships the same, traditional GCNs miss out on the rich information encoded in the different types of edges within a text graph, meaning they overlook the inherent variety in how words interact. 
Relational Graph Convolutional Networks (RGCNs) address this by using different types of edges to capture different relationships. 
However, this expressiveness comes at a computational cost. Nevertheless, heterogeneous graphs provide a more powerful way to represent text, allowing GNNs to adapt their message-passing based on the semantics of different relationships. 
This richer representation often leads to improved performance in text classification tasks.

In this paper, we propose leveraging the ability of Relational Graph Convolutional Networks (RGCNs) to understand relational information for sentiment analysis tasks. 
By incorporating pre-trained language models such as BERT and RoBERTa into the RGCN framework, we aim to enhance the model's ability to extract meaningful sentiment-related features from documents. 
To demonstrate the effectiveness of our approach, we conduct experiments on two diverse datasets: the English-language Amazon reviews dataset and the Persian-language Digikala reviews dataset. 
Through comparison with existing methods, we showcase the superior performance of our RGCN-based approach in capturing relational information.
The remainder of this paper is organized as follows. 
Section 2 provides a brief overview of related work in sentiment analysis and GCNs. 
Section 3 details our proposed method, including the RGCN architecture and training process. 
Section 4 presents the experimental setup and evaluation results. 
Finally, Section 5 concludes the paper and outlines potential future directions. 
Codes are available here.\footnote{\url{https://github.com/agmlabaut/Relational-Graph-Convolutional-Networks-RGCN-/}}

\section{Related Work}
\label{sec:headings}

In recent years, there has been significant research interest in leveraging graph-based models for text classification tasks, aiming to capture the relational dependencies and semantic associations present in textual data. 
Unlike some applications with explicit graph structures including constituency\footnote{The constituency graph is a widely used static graph that can capture phrase-based syntactic relations in a sentence.} or dependency\footnote{A dependency graph is a directed graph representing dependencies of several objects towards each other.} graphs \cite{tang2020integration}, 
knowledge graphs \cite{marin2014learning, ostendorff2019enriching}, social networks \cite{dai2022graph} without constructing graph structure and defining corresponding nodes and edges, text-specific graphs are implicit, which means we need to define
a new graph structure for a specific task such as designing a word-word or word-document co-occurrence graph. 

Two main approaches based on graph construction are corpus-level graphs and document-level graphs.
Corpus-level graph methods encompass the entire collection of text documents, uncovering patterns in word usage across the whole dataset. 
On the other hand, in document-level graphs, the focus is on the internal structure of a single document, capturing how concepts and ideas connect within that specific text.

One notable approach in this domain was Text-GCN \cite{yao2019graph}, which built a corpus-level graph with training document nodes, test document nodes, and word nodes to capture semantic relationships between words and documents. 
A two-layer GCN was applied to the graph, and the dimension of the second layer output equals the number of classes in the dataset. 
TextGCN was the first work that treated a text classification task as a node classification problem by constructing a corpus-level graph and has inspired many following works.

Wu et al. (2019) \cite{wu2019simplifying} proposed Simple Graph Convolution (SGC) to address the computational complexity of Graph Convolutional Networks (GCNs). 
They achieved this by removing the non-linear activation function within GCN layers, resulting in a single linear transformation with comparable or even better performance on various tasks. 
Zhu and Koniusz (2020) \cite{zhu2020simple} proposed Simple Spectral Graph Convolution (S2GC) which included self-loops using Markov Diffusion Kernel to solve the oversmoothing issues in GCN. 
Other than using the sum of each GCN layer in S2GC, the NMGC model which was proposed by Lei et al. (2021) \cite{lei2021multihop} applied min pooling using the Multi-hop neighbor Information Fusion (MIF) operator to address over-smoothing problems. 
Zhang and Zhang (2020) \cite{zhang2020text} introduced TG-Transformer (Text Graph Transformer) which adopted two sets of weights for document nodes and word nodes respectively to introduce heterogeneity into the TextGCN graph.
Lin et al. (2022) \cite{lin2021bertgcn} proposed BertGCN, which aimed to combine the strengths of BERT (Devlin et al., 2018) \cite{devlin2018bert} and TextGCN. 
BertGCN replaced the document node initialization with the BERT's "CLS" output obtained in each epoch and replaced the word input vector with zeros.
Instead of constructing a single corpus-level graph, TensorGCN which was proposed by Liu et al. \cite{liu2020tensor} built three independent graphs: Semantic-based graph, Syntactic-based graph, and Sequential-based graph to incorporate semantic, syntactic, and sequential information respectively and combined them into a tensor graph. 
To fully utilize the corpus information and analyze rich relational information of the graph, Wang et al. (2022) \cite{wang2022me} proposed ME-GCN (Multi-dimensional Edge-Embedded GCN) and built a graph with multi-dimensional word-word, word-document and document-document edges.

Various works have been done to make TextGCN Inductive. 
Ragesh et al. (2021) \cite{ragesh2021hetegcn} optimized TextGCN with HeteGCN (Heterogeneous GCN) by decomposing the original undirected graph into several directed subgraphs. 
Wang et al. (2022) \cite{wang2022induct} aimed to extend the transductive TextGCN into an inductive model with InducT-GCN (Inductive Text GCN). 
Xie et al. (2021) \cite{xie2021inductive} adopted a Variational Graph Auto-Encoder on the latent topic of each document with T-VGAE (Topic Variational Graph Auto-Encoder) to enable inductive learning.

Schlichtkrull et al. (2017) \cite{schlichtkrull2018modeling} introduced a powerful approach called Relational Graph Convolutional Networks (R-GCNs) for modeling relational data. 
Their model effectively learns representations for nodes in a graph by considering not only the node features themselves but also the relationships between nodes. 
In this work, we leverage this concept by employing R-GCNs to use it on sentiment analysis.

BERT \cite{devlin2018bert}, a powerful NLP model by Google, created word representations in sentences. This BERT vector captured a word's meaning within its context. 
Trained on massive datasets like English Wikipedia, BERT assigned a unique code to each word. 
To ensure consistent input lengths for neural networks, sentences were padded with zeros after a fixed length was defined. 
RoBERTa \cite{liu2019roberta}, a more robust Bert, building on BERT's architecture, improved its performance through various refinements. 
Notably, RoBERTa utilized longer pre-training with bigger batches and removed unnecessary pre-training tasks.
ParsBERT \cite{farahani2021parsbert} was a transformer-based pre-trained language model, similar to BERT, that has been specifically trained on a large corpus of formal and informal Persian text data (over 3.9 million documents) for Persian Natural Language Processing (NLP) tasks. 
ParsBERT was trained on a large corpus of Persian text collected from Persian-language websites. 
This pre-training process allowed ParsBERT to learn the text representation vectors of words in Persian text, which can then be fine-tuned for a wide range of NLP tasks such as text classification, named entity recognition, and question answering in Persian.

\section{Proposed Method}
\label{sec:others}

\begin{figure}[htp]
  \centering
  \includegraphics[width=10cm]{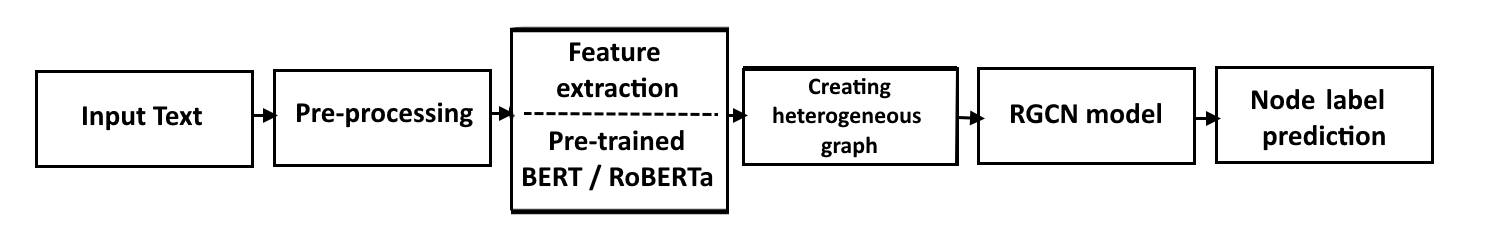}
  \caption{Overview of the proposed method}
  \label{fig:fig4}
\end{figure}

In this section, we propose our method for sentiment analysis which is to construct the heterogeneous graph, calculate the feature vectors of the nodes using the pre-trained BERT and Roberta models, and then feed them into the Relational graph neural network and predict the node labels. 
We initialize node embeddings with pre-trained BERT representations and utilize RGCNs for node classification. 
By using the BERT models, our model benefits from capitalizing on pre-trained BERT, which leverages vast amounts of unlabeled data to capture rich semantic information for text elements.
Figure \ref{fig:fig4} shows the overview of our proposed method.

\subsection{Text Pre-processing}
Our proposed method incorporates text pre-processing to prepare the text data for graph construction. 
This pre-processing includes text normalization (lower casing, removing punctuation and spell checking), removing numbers and extraneous content (URLs, HTML tags), using chat word conversion, and simplifying the text by removing emojis and low-frequency words. 
Additionally, we handle abbreviations and remove stop words and rare words to focus on the core meaning. 
Finally, we tokenize the text and perform lemmatization to ensure consistent word representation. 
Specifically, for the dataset in Persian, we used normalization, punctuation removal, unnecessary word removal, tokenization, number removal, and lemmatization.

\subsection{Graph Construction}
\subsubsection{Heterogeneous graph}

A heterogeneous graph is a more flexible way to represent networks where data can come in various forms \cite{wang2022survey}. 
Unlike a homogeneous graph, a heterogeneous graph allows for different types of nodes and edges. 
Imagine a social media network where you can have users, posts, and comments. 
A standard graph would just represent them as nodes and connections between them. 
But a heterogeneous graph can differentiate between a user node and a post node, and also distinguish between a "likes" edge and a "comments on" edge.
Formally, a heterogeneous graph is denoted by $G = (V, E, \tau, \phi)$, where $V$ is the set of nodes, and the type of node for node $v$ is denoted as $\tau(v)$. 
The set of edges is $E$, and the type of edge for edge $(u, v)$ is denoted by $\phi(u, v)$. 
Additionally, we can also use an ordered triple $r (u, v) = (\tau (u), \phi (u, v), \tau (v))$ to represent relationships in a heterogeneous graph.
Here, the aim is to construct a directed and weighted heterogeneous graph that contains a good representation of the relationships between the nodes in the dataset.

\subsubsection{Creating edges}

The edges between the nodes are created based on the occurrence of a word in the document, the co-occurrence of words in the entire corpus, and the similarity between two documents. 
Therefore, in our heterogeneous graph, we have three types of edges: word-word edges, word-document edges, and document-document edges. 
For each of these edge types, we need to define a weighting metric that captures the strength or importance of the relationship between the connected entities.

\begin{itemize}

\item For calculating the weight of the link between word-word nodes, we use the point-wise mutual information (PMI) method \cite{church1990word}, which is a popular metric for calculating the weight between two-word nodes. 
To calculate the weight of the link between two words $i$ and $j$, PMI is defined in the formula \ref{eq:1}.

\begin{equation} \label{eq:1}
PMI(i. j) = \log(\frac {p(i. j)} {p(i)p(j)})
\end{equation}
\begin{equation} \label{eq:2}
P(i. j) = \frac {\#W(i. j)} {\#W}
\end{equation}
\begin{equation} \label{eq:3}
P(i) = \frac {\#W(i)} {\#W}
\end{equation}

In formula \ref{eq:3}, the number of sliding windows in the entire dataset that contain the word $i$ is denoted by $\#W(i)$. 
Also in formula \ref{eq:2}, $\#W(i, j)$ is the number of sliding windows that contain word $i$ and word $j$. 
The $\#W$ is the total number of sliding windows in the entire dataset. 
Positive PMI values indicate a high semantic relationship between words in the dataset, while negative PMI indicates the absence of a semantic relationship or a weak semantic relationship between two words, therefore, we only add edges with positive PMIs to the graph.
The edges are connected in a bidirectional manner here.

\item For document-document nodes, we use the Jaccard \cite{jaccard1901etude} weighting metric which is used to calculate the similarity between two documents A and B, and is obtained using formula \ref{eq:4}:

\begin{equation} \label{eq:4}
J =  \frac{|A \cap B|}{|A \cup B|}  = \frac{|A \cap B|}{(|A| + |B| - |A \cap B|)} 
\end{equation}

If the two sets are completely equal, then $J = 1$, if they have no common elements, then J = 0; if they have some common elements, then $0 \leq J(A, B) \leq 1$. 
Also, the edges are bidirectional here.

\item For word-document edges we use the Term Frequency-Inverse Document Frequency (TF-IDF) \cite{salton1983introduction} weighting metric. To obtain the TF-IDF score, we need to calculate each of these two terms separately and multiply the results together. The resulting score will show us the weighted frequency of the keyword. The formula is given in \ref{eq:5}:

\begin{equation} \label{eq:5}
TF-IDF = tf_{x,y}  * IDF = tf_{x,y} * \log(\frac{N}{df_x})
\end{equation}

The $N$ is the total number of words in the content, $tf_{x,y}$ is the number of times the word $x$ appears in the document $y$ divided by the number of words in the document $y$, $df_x$ is the number of documents that contain the word $x$ and IDF is the logarithm of the number of total content divided by the content that contains the word $x$.

To obtain the co-occurrence information of keywords, we use a sliding window with a fixed size on all text documents in the dataset to collect co-occurrence statistics.
Additionally, the direction of edges is from documents to words.

\end{itemize}

\begin{figure}[htp]
  \centering
  \includegraphics[width=10cm]{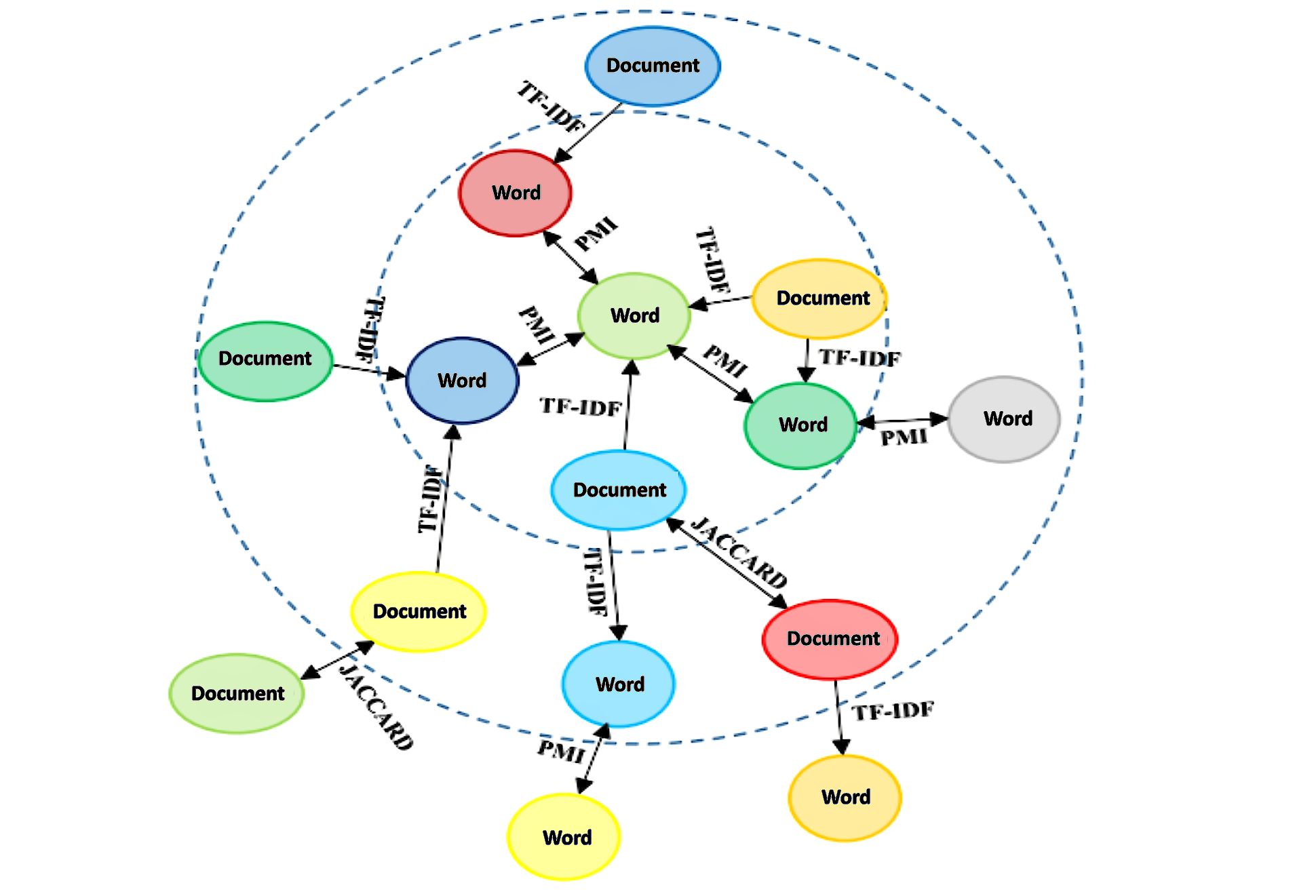}
  \caption{Overview of the constructed graph - The constructed graph is a directed and weighted heterogeneous graph with two types of nodes (words and documents) and three types of edges ( word-word, word-document, and document-document )}
  \label{fig:fig1}
\end{figure}

Figure \ref{fig:fig1} shows an overview of the constructed graph using the mentioned three metrics.

\subsection{Pre-trained language models}

This work leverages BERT (RoBERTa) to generate contextualized representations for both document and word nodes.
After constructing the graph, we first create an initial representation vector, using the pre-trained BERT model and the pre-trained RoBERTa model for each of the nodes (documents and words) in the graph. 
Then, these representation vectors are fed into the RGCN as the initial features of the nodes.
For each document, $D_{d}$, we process it through the pre-trained BERT (RoBERTa) model. This can be formulated as \ref{eq:9}.

\begin{equation} \label{eq:9}
B_{D_{d}} = BERT/RoBERTa(D_{d})
\end{equation}

\begin{equation} \label{eq:10}
H_{node_{D_{d}}} = B_{D_{d}}^{[cls]}
\end{equation}

\begin{equation} \label{eq:11}
H_{node_{W_{m}}} = min_{d \in D_{W_{m}}} (B_{D_{d}}^{W_{m}})
\end{equation}
 
For example, we give the sentence “John feels happy” to BERT (RoBERTa) and the output would be $ B_{D_{d}}^{[CLS]} B_{D_{d}}^{John}B_{D_{d}}^{feels}B_{D_{d}}^{happy}B_{D_{d}}^{[SEP]}$. 
Using formula \ref{eq:10}, we consider the representation vector "$[CLS]$" as the representation vector of the document $D_{d}$. 
This "$[CLS]$" token, a characteristic of BERT and RoBERTa, is known to encapsulate a comprehensive representation of the entire input sequence. Consequently, $H_{node_{D_{d}}}$ serves as the final document node representation for the RGCN.

The formula \ref{eq:11} defines the representation vector for a word node $W_{m}$. 
We first consider all documents $D_{W_{m}}$ that contain the word $W_{m}$. 
Subsequently, for each document $D_{d}$ in this set, we extract the specific embedding vector $B_{D_{d}}^{W_{m}}$ corresponding to the word $W_{m}$ within the document embedding $B_{D_{d}}$.
Finally, we employ min-pooling to select the minimum vector across all documents containing $W_{m}$.
This min-pooled vector, $H_{node_{W_{m}}}$, captures the most prominent contextual representation of the word across different document occurrences.

\subsection{RGCN framework}

RGCN is a convolution operation that performs message passing on multi-relational graphs. 
The RGCN method is used to learn the representation vectors of the nodes in the graph. 
The difference between the RGCN method and the GCN method is that the GCN method operates on graphs with one type of edge, while the RGCN method operates on graphs with multiple types of edges and considers a different processing channel for each type of edge. 
We explain how graph convolutional networks operate on directed graphs and how they can be extended to relational graphs. 
We describe message passing in terms of matrix multiplication and explain the intuition behind this operation. 
This model is a generalized version of GCN that operates on graphs in large-scale multi-relational data. 
Related methods, such as graph neural networks, can be understood as special cases of the message-passing framework.

\begin{equation} \label{eq:6}
h_i^{(l+1)} = \sigma \left( \sum_{m \in M_i} g_m(h_i^{(l)}, h_j^{(l)}) \right)
\end{equation}

In formula \ref{eq:6}, $h_i^{(l)} \in R^{d(l)}$ is the hidden state of node $v_i$ at layer $l$ of the neural network, where $d^{(l)}$ is the dimension of this layer. 
The received messages are aggregated using the function $g_m(\cdot,\cdot)$ and passed through an element-wise activation function $\sigma(\cdot)$, such as $ReLU(0) = max(0,\cdot)$. 
The $M_i$ denotes the set of incoming messages for node $v_i$ and is often chosen to be the same as the set of incoming edges. 
The $g_m(\cdot,\cdot)$ is typically chosen to be either a neural network function (message-specific) or simply a linear transformation $g_m(h_i,h_j) = W h_j$ with a weight matrix $W$. 
This is essentially a transformation of the message before it is passed on. 
This type of transformation is very effective in aggregating and encoding local features of the graph structure's neighbors and has led to significant advances in areas such as graph classification and graph-based semi-supervised learning.

Motivated by these architectures, we define the following simple propagation model for computing the forward-pass update of an entity or node denoted by $v_i$ in a multi-relational (directed and labeled) graph.

\begin{equation} \label{eq:7}
h_i^{(l+1)} = \sigma \left( \sum_{r \in R} \sum_{j \in N_i^{r}}  \frac{1}{c_{i,r}} W_r^{(l)} h_j^{(l)} + W_0^{(l)} h_i^{(l)} \right)
\end{equation}

In formula \ref{eq:7}, $N_i^{r}$ denotes the index set of node i's neighbors under relation $r \in R$. 
The $c_{i,r}$ is a problem-specific normalization constant that can be learned or chosen beforehand (e.g., $c_{i,r} = |N_i^{r}|$).
Intuitively, the model aggregates the transformed feature vectors of neighboring nodes via a normalized summation.

\begin{figure}[htp]
  \centering
  \includegraphics[width=10cm]{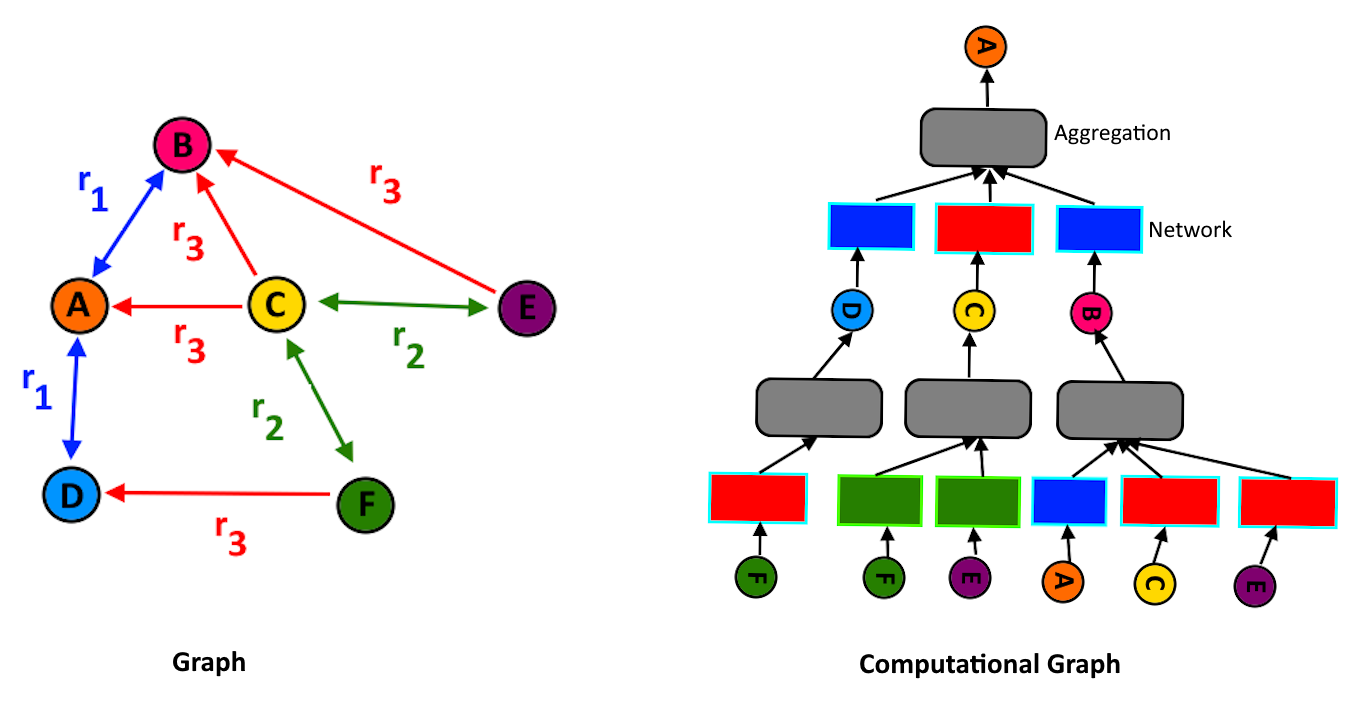}
  \caption{Overview of RGCN and its corresponding computational graph for the example node "A".}
  \label{fig:fig2}
\end{figure}

Different from conventional GCNs, in this model, relation-specific transformations, represented by the relation weight matrix $W_r^{(l)}$, are introduced depending on the type and direction of an edge. 
As shown in Figure \ref{fig:fig2}, three weight matrices are shown with three different colors.

To ensure that the representation of a node in layer $l+1$ can also be aware of the corresponding representation of the same node in layer $l$, a self-loop of a special relation type is added to each node. 
Note that instead of simple linear message transformations, more flexible functions (of course, with computational cost considerations) such as multilayer neural networks can be selected. 
The formula \ref{eq:7} is a layer update of the neural network in parallel for each node in the graph. 
In practice, the formula can be efficiently implemented using sparse matrix multiplication to avoid explicit summation over neighbors. 
Multiple layers can be stacked on top of each other to allow for dependencies over multiple relational steps. 
This model is referred to as a Relational Graph Convolutional Network (RGCN).

\begin{figure}[htp]
  \centering
  \includegraphics[width=10cm]{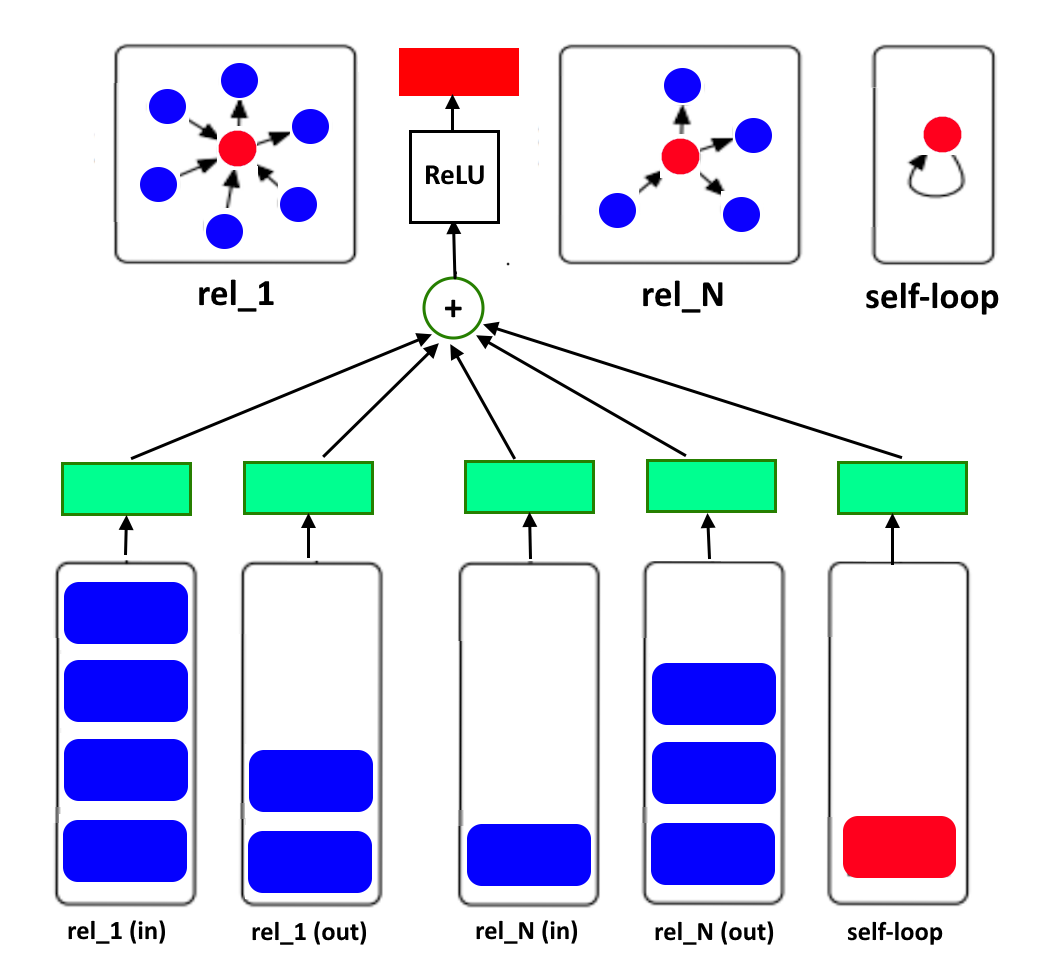}
  \caption{The computation view for updating a single graph node in the RGCN model} 
  \label{fig:fig3}
\end{figure}

The computational graph for updating a node in the RGCN model is shown in figure \ref{fig:fig3}. 
The d-dimensional feature vectors of the neighboring nodes (in blue) are first aggregated and then transformed separately for each relation type (both for incoming and outgoing edges). 
The resulting representation (green) is aggregated in a normalized summation and passed through an activation function. 
This update can be computed for each node in parallel with parameters shared across the entire graph. 
The overall RGCN model is thus to stack $l$ layers as defined in the above formula, with the output of the previous layer being the input to the next layer. 
If no node features exist, the input to the first layer can be chosen as a one-hot vector for each node in the graph. 
A problem with directly applying the above equation is the rapid growth of the number of parameters, especially for data with a large number of edges. 
In order to reduce the number of parameters of the model and prevent overfitting, the use of basis decomposition has been proposed. With the basis decomposition, each $W_r^{(l)}$ is defined as formula \ref{eq:8}.

\begin{equation} \label{eq:8}
W_r^{(l)} = \sum_{b=1}^{B} a_{rb}^{(l)} V_{b}^{(l)}
\end{equation}

i.e. as a linear combination of basis transformations $V_{b}^{(l)} \in R^{d^{(l+1)} * d^{l}}$ with coefficients $a_{rb}^{(l)}$ such that only the coefficients depend on r.

\subsection{Details of our proposed method}

\subsubsection{Training settings}

There are two settings of learning in GNNs: inductive and transductive \cite{ciano2021inductive}.
Inductive learning is the same thing as what we usually know as supervised learning. 
We build and train a machine learning model based on a labeled training dataset that we already have (just training nodes and not test nodes). 
Then, we use this trained model to predict the labels of a test dataset that we have never seen before \cite{xie2021inductive, wang2022induct, ragesh2021hetegcn}.

On the other hand, in transductive learning, all data, both the training and test datasets, are observed in advance. 
They learn from the observed training dataset including only the labels of the train data and then predict the labels of the test dataset. 
Even if we do not know the labels of the test dataset, we can use the patterns and additional information in this data during the learning process \cite{zhang2020text, schlichtkrull2018modeling}.

Our model is trained in a transductive manner because, according to the transductive logic, it uses the entire graph structure to obtain embeddings, i.e., the connections affect message passing. 
However, training is done using the labels of the separate sections. 
The graph is constructed at the text corpus level, i.e., a graph is created over the entire text data, and the word embedding is obtained from the min pooling of the embeddings of the documents it contains. 
Therefore, the document that has labels is included in all three sections: training, validation, and testing. And the word must also exist in order to preserve the complete connections. 
Finally, the entire graph structure plays a role in training the model.

\subsubsection{The model}

Our proposed RGCN network consists of two layers, which allows information exchange between nodes that are at most two hops (neighbors and neighbors of neighbors) apart. 
The activation function is the ReLU function, and the optimizer function is the ADAM function. 
The first layer of the RGCN model is fed with a feature vector of each node which is acquired from BERT or RoBERTa. 
Based on our obtained results, a two-layer RGCN performs better than a one-layer RGCN, while increasing the number of layers does not improve the performance of the model. 
The size of the representation vectors of the second layer is the same as the size of our class set, which is updated with RGCN based on the graph structure. 
The final representation vectors obtained for each of the nodes of the document are considered as the output of the RGCN, which are passed through a SoftMax classifier to perform the prediction.

By constructing a large heterogeneous text graph containing word nodes and document nodes, we can explicitly model both word co-occurrences and easily apply graph complexities. 
The number of nodes in the text graph is equal to the number of documents (corpus size) plus the number of distinct words in the set (vocabulary size).

We define the edge between document and word nodes as directed (from document to word) and the edge between two-word nodes or two document nodes as bidirectional. 
A heterogeneous graph is defined with the structure $(V, E, \tau, \phi) = G$, where:

\begin{itemize}

\item Node type $\tau(v)$ is defined as word or document.
\item Edge type $\phi (u, v)$ is defined as co-occurrence, similarity, or frequency.
\item The relationship type is expressed as an ordered triple for R:
(Word, Co-occurrence, Word), (document, Similarity, document), (document, Frequency, Word)

\end{itemize}

Since the rule of message passing in a directed graph is such that messages are only sent in the direction of the edge, the problem is that for a triple $<s. r. o>$ message is sent from s to o, but not from o to s. 
For example, for the triplet <Amsterdam.Located in. The Netherlands> Amsterdam needs to be updated with information from the Netherlands and the Netherlands needs to be updated with information from Amsterdam, defining different directions has different meanings in modeling.
In order for the model to be able to send messages in both directions, the graph inside the RGCN layer is corrected by adding reverse edges, so that for each existing edge $<s. r. o>$, a new edge $<s. \acute{r}. o>$ is added where $\acute{r}$ is a new relation that represents inverse of r.

As we know, in the simple implementation of RGCN or GCN, the output representation for node $i$ does not preserve any of the input representation information of each node. 
To allow such information to be preserved, a self-loop $<s. r_{s}. s>$ is added to each node.

For each relation in the graph, a sparse adjacency matrix $A_{r}$ is formed. 
Reverse edges are only considered for bidirectional relations, and the adjacency matrix becomes symmetric considering self-loops. 
Therefore, the size of the adjacency matrix is determined based on the number of nodes in the source and destination node types of the relation. 
This ensures that the adjacency matrix has the correct dimensions to represent that particular relation in the heterogeneous graph. 
Considering the above, the update of the embedding of a node in our designed model is as follows:

\begin{equation} \label{eq:12}
h_i^{(l+1)} = ReLU \left( \sum_{r \in R} \sum_{j \in N_i^{r}}  \frac{1}{N_i^{r}} W_r^{(l)} h_j^{(l)} + W_0^{(l)} h_i^{(l)} \right)
\end{equation}

If we consider formula \ref{eq:12} in matrix form for each node, we get formulas \ref{eq:13} and \ref{eq:14}:

\begin{equation} \label{eq:13}
H^{(k)} = [h_{1}^{(k)} ... h_{|V|}^{(k)}]^{T}
\end{equation}
\begin{equation} \label{eq:14}
H^{(l+1)} = ReLU \left( \sum_{r=1}^{R} A_{r} H^{l} W_{r}^{l} + H^{l} W_{0}^{1} \right)
\end{equation}

\section{Experiments}

\subsection{Datasets}

\subsubsection{The Amazon dataset}

The dataset used in this research is the Amazon dataset\footnote{\url{ https://jmcauley.ucsd.edu/data/amazon/}}, which was collected from the company's website between May 1996 and July 2014. 
The Amazon website sells a wide variety of products from different categories, such as electronics, books, clothing, and more. 
The website allows users to rate products and write reviews. 
All of the information mentioned above is collected in text files in the Amazon dataset.
This dataset consists of 142.8 million samples. 
Each sample in this file represents a user's review of a product. 
Each sample includes information such as the user ID, the item ID, the user's rating of the item (a number between 1 and 5), the user's review of the item (in text form), and the time the user submitted the review.
In this paper, we use the core5-version of the user reviews file (each user or item has at least 5 reviews), on the grocery shopping category (including food, vegetables, prepared meals, etc.), which contains 9982 users, 8682 items, and 151,254 ratings and reviews (22 people without text reviews). 
The specifications of this dataset after pre-processing are shown in Table \ref{tab:table1}.
We also used a 2-class version of this dataset. This dataset has two labels: label 1 (combination of classes 1 and 2 from 5-core version) and label 2 (combination of classes 4 and 5 from 5-core version)  

\begin{figure}[htp]
  \centering
  \includegraphics[width=15cm]{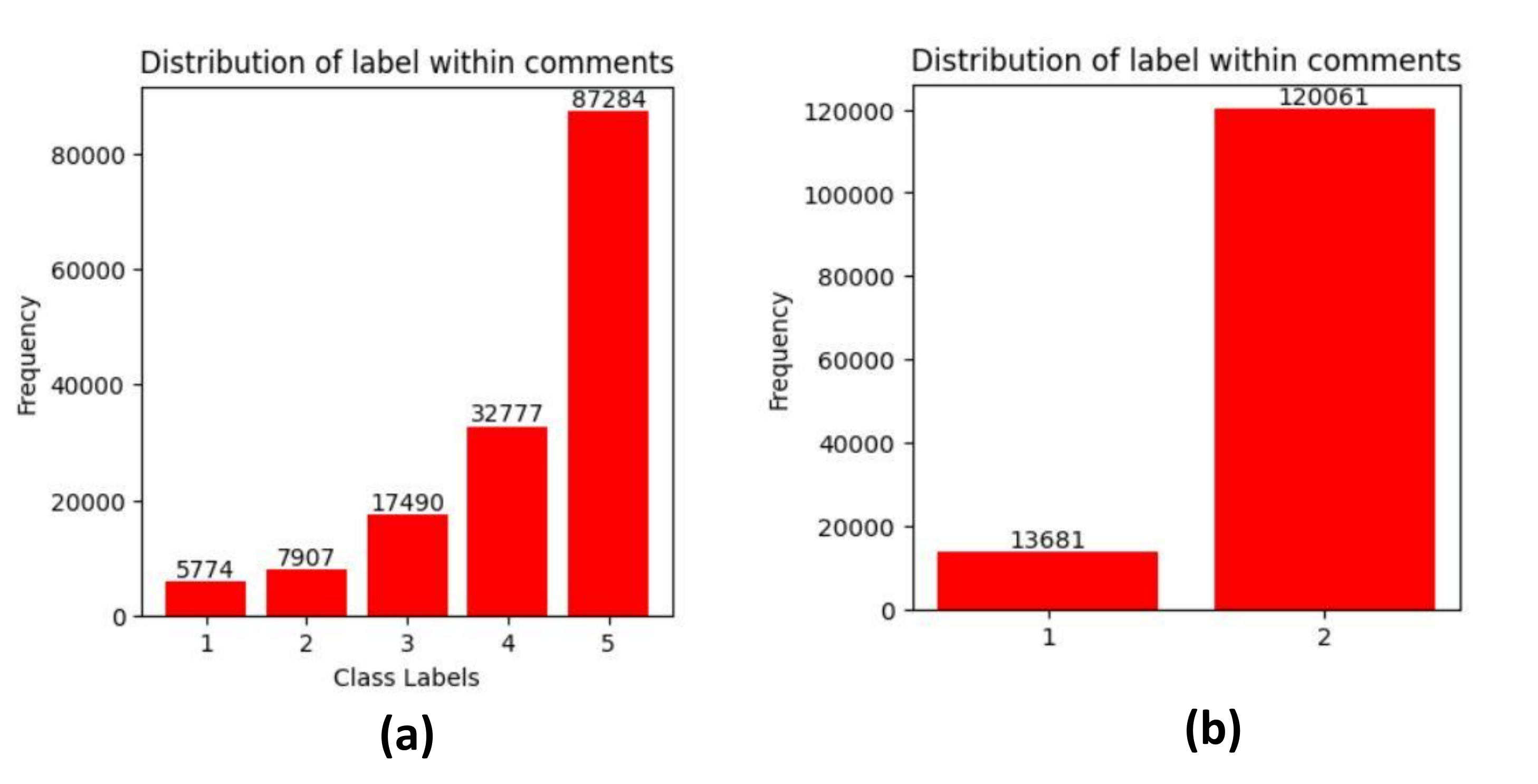}
  \caption{(a) distribution of labels within comments for the dataset with 5 classes (b) distribution of labels within comments for the dataset with 2 classes} 
  \label{fig:fig5}
\end{figure}

Figure \ref{fig:fig5} shows the 5-class and 2-class Amazon datasets before balancing. It shows that before balancing the data, in 2-class version, the number of data with label 2 (combination of classes 4 and 5) is more than the label 1 (combination of classes 1 and 2). 
We can say that label 2 is the majority class and the other class is the minority class.
In addition, in the 5-class version, the class with label 5 is the majority class and the class with label 1 is the minority class. 
We increased the data of classes with labels 1, 2, and 3 to the number of classes with label 4 with the oversampling method, and we reduced the data of classes with label 5 to the number of data of classes with label 4 with the undersampling method. 
Therefore, to balance the datasets, we used a combination of reducing the majority class and increasing the minority class in both versions. 
After balancing 5-class Amazon dataset has 32777 comments in each label and the 2-class Amazon dataset has 13681.

The problem with a model trained on imbalanced data is that the model learns that it can achieve high accuracy even if it does not predict the minority class accurately. 
This can be a problem when applying the model to a real-world problem, where it is more important to predict the minority class accurately. 
Balancing a dataset makes it easier to train a model, as it helps to prevent the model from becoming biased towards one class. 
In other words, the model will no longer favor the majority class, just because it has more data.

\begin{figure}[htp]
  \centering
  \includegraphics[width=20cm]{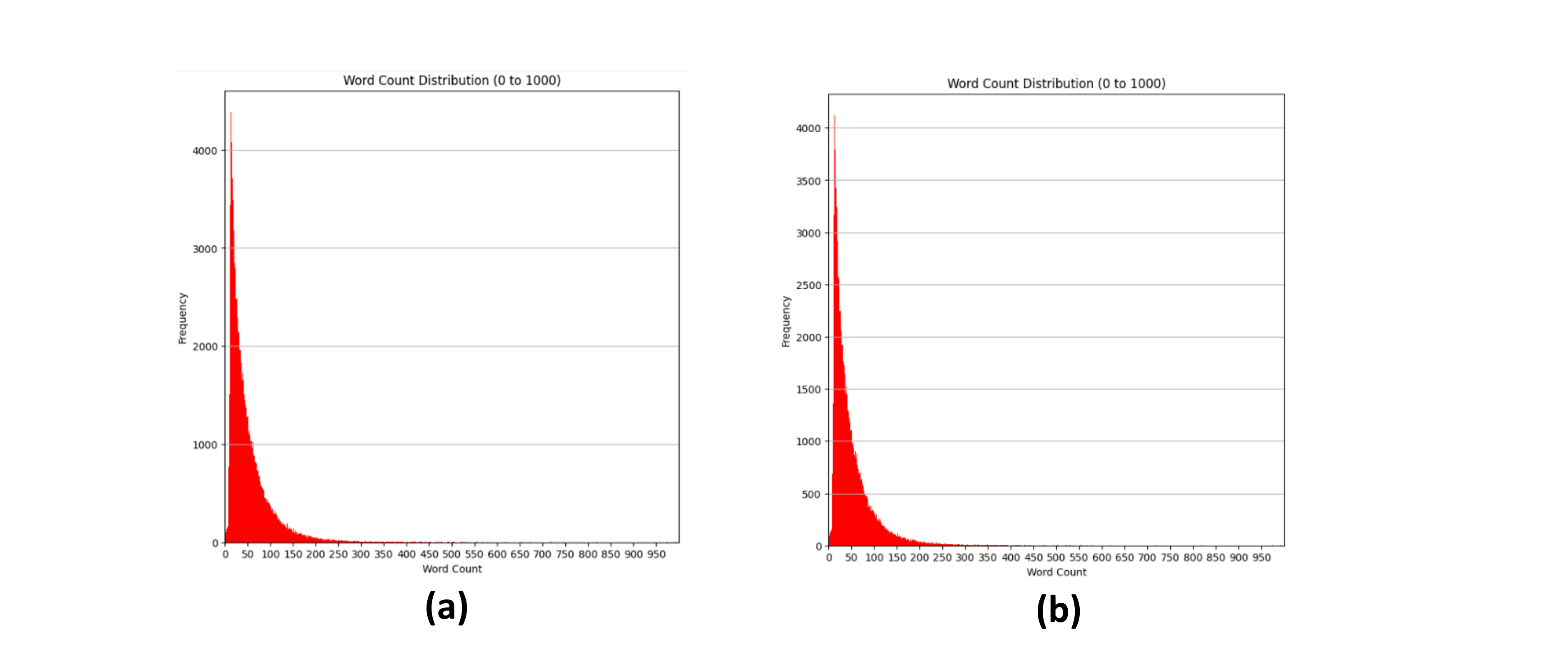}
  \caption{(a) distribution of word count within comments for the dataset with 5 classes (b) distribution of word count within comments for the dataset with 2 classes} 
  \label{fig:fig6}
\end{figure}

In Figures \ref{fig:fig6} (a) and \ref{fig:fig6} (b), we present the distribution of the number of words in the 5-class and 2-class datasets.

\subsubsection{The Digikala dataset}

We also evaluated the proposed model on the Digikala dataset which is in Persian. 
This dataset consists of 100,000 rows and 12 different columns, including user reviews, product pros and cons, number of likes and dislikes, product ID, and more. 
We evaluated the performance of the proposed model in two cases:

\begin{itemize}
    \item In the case where our data has two classes and has two labels "Recommended" and "Not Recommended",
    \item In the case where the data has three classes and has three labels "Recommended", "Not Recommended" and "No Opinion".
\end{itemize}

The specifications of this dataset after preprocessing are shown in Table \ref{tab:table3}.

\begin{table}
    \centering
    \caption{Amazon dataset statistics}
    \label{tab:table1}
    \begin{tabular}{ccccccc}
    \toprule
    \textbf{tag} & \textbf{1} & \textbf{2}& \textbf{3}& \textbf{4}& \textbf{5}& \textbf{Total} \\
    \midrule
    \textbf{number of Sentences}  & 5774  & 7907 & 17490 & 32777 & 87284 & 151232\\
    \bottomrule
    \end{tabular}
\end{table}

\begin{table}
    \centering
    \caption{Digikala dataset statistics}
    \label{tab:table3}
    \begin{tabular}{ccccc}
    \toprule
    \textbf{} & \textbf{Not Recommended} & \textbf{No Opinion}& \textbf{Recommended}& \textbf{Total} \\
    \textbf{Tag} & \textbf{-1} & \textbf{0}& \textbf{1}& \textbf{-} \\
    \midrule
    \textbf{Number of Sentences}  & 16098  & 10528 & 36960 & 63586\\
    \bottomrule
    \end{tabular}
\end{table}

\subsection{Experiment setup}

The proposed method in this research is implemented using the Python development environment. 
The Geometric PyTorch library\footnote{\url{https://pytorch-geometric.readthedocs.io/en/latest}}, which is a PyTorch-based library designed for implementing graph neural networks, was used to implement the graph convolutional and graph relational convolutional networks.
The Hazm\footnote{\url{https://github.com/roshan-research/hazm}} library was used for natural language processing in Persian, and the spaCy\footnote{\url{https://spacy.io}} library was used for English.

\subsection{Evaluation metrics}

To evaluate the performance of the proposed models, we use the cross-entropy loss function, the accuracy metric, and the F1 score. 
Cross entropy is a concept that is commonly used in statistics and machine learning, often as a loss function for measuring the dissimilarity between the predicted probability distribution and the actual distribution of a classification problem. 
The intuition behind cross-entropy is that it measures how much the predicted probabilities match the real probabilities. 
The cross-entropy between the real distribution P and the predicted distribution Q is calculated as:
$cross entropy(P, Q) = - \sum_{i} p_{i} * \log (q_{i})$

Accuracy and F1 score are two metrics that are commonly used to evaluate text classification methods and are calculated according to the following formulas:
Accuracy = (TP + TN) / (TP + FP + FN + TN),
Precision = TP / (TP + FP),
Recall = TP / (TP + FN),
F1-score = 2 * (Precision * Recall) / (Precision + Recall),
Where:
TN: Represents the number of records that the model correctly identified as negative and labeled as negative.
TP: Represents the number of records that the model correctly identified as positive and labeled as positive.
FP: Represents the number of records that the model incorrectly identified as negative but were labeled as positive.
FN: Represents the number of records that the model incorrectly identified as positive but were labeled as negative.

\subsection{Results}

\subsubsection{Results on Amazon dataset}

We evaluate the performance of the proposed model in two cases:
\begin{itemize}
    \item In the case where our data has two classes (excluding class with label 3), the first class includes reviews with labels 1 and 2, and the second class includes reviews with labels 4 and 5.
    \item In the case where our data has five classes, they have labels 1, 2, 3, 4, and 5.
\end{itemize}

In each of these two cases, the performance of the model is evaluated when the data is imbalanced and when the data is balanced. The results of this comparison are presented in Table \ref{tab:table2}.

\begin{table}[htbp]
\centering
\caption{Amazon dataset Results}
\label{tab:table2}
\begin{tabular}{|c|ccc|ccc|}
\hline
\multirow{3}{*}{Model} & \multicolumn{3}{c|}{2 class} & \multicolumn{3}{c|}{5 class} \\ \cline{2-7} 
 & \multicolumn{1}{c|}{Balanced} & \multicolumn{2}{c|}{Imbalance} & \multicolumn{1}{c|}{Balanced} & \multicolumn{2}{c|}{Imbalance} \\ \cline{2-7} 
 & \multicolumn{1}{c|}{Accuracy} & \multicolumn{1}{c|}{Accuracy} & F1-score & \multicolumn{1}{c|}{Accuracy} & \multicolumn{1}{c|}{Accuracy} & F1-score \\ \hline
BERT & 64.46 & 85.88 & 71.42 & 42.00 & 56.98 & 50.06 \\
RoBERTa & 65.35 & 87.67 & 72.26 & 43.10 & 57.04 & 51.43 \\
BERT + GCN & 66.48 & 89.60 & 72.36 & 42.37 & 57.09 & 50.17 \\
RoBERTa + GCN & 67.25 & 89.66 & 73.50 & 43.96 & 57.18 & 52.70 \\
BERT + RGCN & 70.53 & 89.75 & 72.54 & 43.13 & 57.25 & 50.28 \\
RoBERTa + RGCN & 70.59 & 89.82 & 73.92 & 44.28 & 57.28 & 52.83 \\ \hline
\end{tabular}
\end{table}

In table \ref{tab:table2}, We can see that all the percentages in RoBERTa is higher that the numbers gained in the BERT model, for example, accuracy in balanced dataset is 65.35\% in RoBERTa but it is 64.46\% in BERT in the 2-class version and 43.10\% to 42\% in 5-class version. Transformers for Natural Language Processing beyond BERT refer to advanced transformer-based models such as RoBERTa that were created to improve upon the limitations of BERT in natural language processing tasks. 
These models leverage the transformer architecture's ability to handle long-range dependencies and context-sensitive embeddings to offer improved performance on a variety of natural language processing tasks. 
BERT revolutionized the field of natural language processing by introducing a bidirectional transformer-based model that could understand a word's context based on its entire surroundings (both left and right of the word). 
However, subsequent models such as RoBERTa were developed to address some of BERT's limitations, such as pretraining incoherence fine-tuning inefficiency, and inability to use the full context of a sentence in the masked language model. 
These changes lead to a significant improvement in its performance over BERT as can be seen from table \ref{tab:table2}. 
RoBERTa, despite its advantages, also comes with a set of challenges. 
Due to its large size and complexity, it requires significant computational resources and time to train. 
Additionally, given its capacity, it can easily overfit on smaller datasets if not fine-tuned properly.

As can be seen in the table \ref{tab:table2}, compared to the performance of BERT and RoBERTa models, in both balanced and imbalanced data cases, when graph neural networks such as RGCN and GCN are used in combination with the above language models, the performance of the models (RGCN/GCN) + BERT increases from BERT and (RGCN/GCN) + RoBERTa from RoBERTa, which is due to the advantages of using graph neural networks, some of which were mentioned above. 
For example in balanced 5-class version of the dataset, the accuracy of the model GCN + RoBERTa is 43.96\%, which is higher than the accuracy of the model GCN + BERT, which is 42.37\%. 
According to the research and results, GCN has a much higher accuracy than other methods, which shows that GCN is a much more standard model in classification and natural language processing problems. 
Each of these methods has its advantages and may be suitable for different scenarios.
GNNs are designed to effectively capture rich semantic relationships and dependencies between nodes in a graph, which enables better understanding and representation of text content. 
Text classification often requires considering the contextual information of words or phrases to make accurate predictions. 
GNNs can aggregate information from neighboring nodes in the graph, which allows them to effectively collect and propagate textual information. 
This allows GNNs to use the local context of each node and make informed decisions about text classification problems. 
On the other hand, one of the challenges of text classification is dealing with inputs of different lengths, such as sentences and phrases with different numbers of words. 
GNNs can naturally handle variable-length inputs using the graph structure. 
GNNs can depict the relationships and dependencies between words or sentences and provide a more robust and flexible approach to text classification. 
GNNs also perform well in modeling long-range dependencies in a graph. 
In text classification, long-range dependencies refer to dependencies that span the entire dataset and text. 
By propagating information throughout the graph, GNNs can capture these long-range dependencies and enable a comprehensive understanding of textual data and its classification. 

Finally, in table \ref{tab:table2}, we can see RoBERTa have more accuracy when used with RGCN than the case when it is used with GCN. 
Best percentages are acquired from RoBERTa + RGCN model, it has 70.59\% and 44.28\% accuracy in balanced versions and 73.92\% and 52.83\% F1 score in imbalanced versions.
Relational Graph Convolutional Networks (RGCNs) offer advantages over traditional Graph Convolutional Networks (GCNs) in scenarios where the relationships between nodes are complex and diverse. 
The relationships between entities (e.g., words, documents, and entities mentioned in the text) can be complex and diverse. 
RGCNs can model such complex relationships more effectively than GCNs. 
For example, in document classification, words may have different types of relationships (such as grammatical dependencies, semantic relations, and structural relations like co-occurrence), and RGCNs can capture and utilize these relationships more fully. 
RGCNs allow for relation-specific message passing, which means that they can define different transformation matrices ($W_r$) based on the type of relationship between nodes. 
As a result, a specific message ($h*W_r$) for that relationship is sent for each type of relationship, enabling the model to distinguish between different types of interactions and leading to representations with finer differences. 
In contrast, GCNs typically apply the same transformation to all edges in the graph, which may not be optimal for capturing the diverse relationships in text documents. 
RGCNs provide a more expressive framework for modeling structured graph data. 
By explicitly incorporating relation-specific parameters, they offer more flexibility in learning representations that are tailored to the specific characteristics of the data and can lead to better performance, especially in tasks where relationships play an important role, such as social networks, and text classification with rich semantic and syntactic dependencies. 
Given that a separate adjacency matrix ($A_r$) is defined for each type of relationship in RGCNs, this adjacency matrix represents not only the presence of edges between nodes but also the type or nature of the relationships. 
As a result, they are more flexible than GCNs and can better generalize to unseen or heterogeneous data. 
By learning separate parameters for different types of relationships, the model becomes more robust to changes in the distribution of the data and increases its ability to adapt to different contexts and domains. 
This helps to solve the challenge of sparsity and leads to improved results and robustness of text classification models.

In summary, RGCNs offer advantages over GCNs in handling complex relationships and capturing diverse interactions in documents, while relation-specific adjacency matrices provide a richer representation of relationships, increasing the capacity of the model and improving its generalization capabilities. 
These characteristics make RGCNs particularly suitable for message passing in text classification tasks.

\subsubsection{Results on Digikala dataset}

\begin{table}[htbp]
\centering
\caption{Digikala dataset Results}
\label{tab:table4}
\begin{tabular}{|c|ccc|ccc|}
\hline
\multirow{3}{*}{Model} & \multicolumn{3}{c|}{2 class} & \multicolumn{3}{c|}{3 class} \\ \cline{2-7} 
 & \multicolumn{1}{c|}{balanced} & \multicolumn{2}{c|}{imbalance} & \multicolumn{1}{c|}{balanced} & \multicolumn{2}{c|}{imbalance} \\ \cline{2-7} 
 & \multicolumn{1}{c|}{accuracy} & \multicolumn{1}{c|}{accuracy} & F1-score & \multicolumn{1}{c|}{accuracy} & \multicolumn{1}{c|}{accuracy} & F1-score \\ \hline
ParsBERT & 68 & 87 & 72 & 57 & 62 & 55 \\
ParsBERT + GCN & 70 & 91.1 & 74 & 58 & 63.9 & 55 \\
ParsBERT + RGCN & 70.36 & 91.17 & 74.15 & 58.29 & 63.94 & 55.11\\ \hline
\end{tabular}
\end{table}

ParsBERT is pre-trained on a large corpus of Persian text, which allows it to capture language features and specific differences in the Persian language.

Table \ref{tab:table4} shows the results and accuracy of the proposed model on the Digikala dataset. As can be seen, the combination of the ParsBERT model with the RGCN model achieved better results than the combination of ParsBERT with GCN. Using ParsBERT and GCN on the balanced two-class dataset with 500 epochs, we achieved an accuracy of 70.36\% on the test data and an accuracy of 71.51\% on the training data. Increasing the number of epochs to 1000 resulted in an accuracy of 70.26\% on the test data and an accuracy of 71.86\% on the training data. It is predicted that in this case, the model has overfitted, and we reached an accuracy of 58.29\% on the test data and an accuracy of 59.74\% on the training data in the balanced three-class dataset with 1000 epochs, considering a learning rate of 0.01 and DROPOUT of 0.5. It is worth noting that the amount of performance improvement of the RGCN + ParsBERT model compared to the GCN + ParsBERT model in Persian is less than the amount of improvement of the RGCN + (BERT/ RoBERTa) models compared to GCN + (BERT/ RoBERTa) in English. Overall, it can be said that English and Persian languages have significant differences in terms of grammar, syntax, and language features, some of which are mentioned below:
\begin{itemize}
    \item In English, words are usually separated by spaces, which makes tokenization relatively simple. In Persian, words are connected and there is no clear space between them, so the tokenization process is more difficult.
    \item In Persian, compound words are often formed by combining several single words.
    Correctly tokenizing these combinations can sometimes be difficult, as the boundaries between the constituent words must be accurately identified.
    \item The complex morphological problem: In Persian is such that Persian words can undergo extensive changes through prefixes, suffixes, and root inflections to indicate different grammatical features. This complexity adds layer of difficulty to tokenization.
    \item Persian words can be ambiguous, meaning that a string of characters can have multiple valid interpretations. Resolving this ambiguity during tokenization and stemming requires a deeper understanding of the language and its context.
    
\end{itemize}

\section{Conclusion and Future work}

\subsection{Conclusion}

In this paper, we considered the classification of English and Persian texts and addressed this issue using graph neural networks, focusing on relational graph convolutional networks. 
We made a heterogeneous text graph from the entire dataset (all documents and all unique words) and constructed the graph by defining three types of relationships (co-occurrence, similarity, and frequency) and weighted edges IDF-TF, PMI, and Jaccard. 
We then used pre-trained models such as BERT and RoBERTa to extract features for each node and used a two-layer relational graph convolutional network to train the model, which improved the results compared to the graph convolutional network.

\subsection{Future work}

Many improvements can be made in this research area. 
Most supervised deep-learning models are trained on large amounts of labeled data. 
In practice, collecting such labels in any new domain is expensive. 
A language model (such as BERT) with fine-tuning for a specific task requires much less labeling than training a model from scratch. 
Therefore, there are opportunities to develop new methods such as zero-shot or few-shot learning based on these language models.

In most text classification methods with GNNs, edges with a fixed value extracted from statistical information of documents are used to construct graphs. 
This approach is applied to GNNs at the corpus level and the document level. 
However, to better investigate the complex relationship between words and documents, it is suggested to use dynamic edges. 
Dynamic edges in GNNs can be learned from various sources such as graph structure, semantic information of documents, or other models.

In addition, a more complex algorithm can be used to calculate the edge between two documents, and it is recommended to filter the word-word and word-document relationships to simplify the graph structure, i.e., identify important edges based on a criterion and only store them.

GNN text classification models perform well at the corpus level. 
These models are mostly transductive, meaning that they only work on the present graph and cannot be applied to new nodes and edges. 
when a new document is added, the graph must be reconstructed from scratch, which is very expensive and impractical for real-world applications. 
Therefore, it is worth considering the inductive learning approach.

Since the layers of RGCN are stacked in such a way that the input of one layer is the output of the previous layer, taking the sum over the relationships causes an accumulation of activations. 
However, for two-layer networks, it does not seem that this event affects the performance of the model. 
For deeper models, taking the average over the relationships instead of the sum may be more appropriate.

\bibliographystyle{unsrt}  
\bibliography{references}  

\end{document}